# BERTologyNavigator: Advanced Question Answering with BERT-based Semantics

Shreya Rajpal[1,2], Ricardo Usbeck[1]

[1] *Universität Hamburg, Hamburg, Germany*
[2] *Vellore Institute of Technology, Vellore, Tamil Nadu, India*

**Abstract**

The development and integration of knowledge graphs and language models has significance in artificial intelligence and natural language processing. In this study, we introduce the BERTologyNavigator- a two-phased system that combines relation extraction techniques and BERT embeddings to navigate the relationships within the DBLP Knowledge Graph (KG). Our approach focuses on extracting one-hop relations and labelled candidate pairs in the first phases. This is followed by employing BERT's CLS embeddings and additional heuristics for relation selection in the second phase. Our system reaches an F1 score of 0.2175 on the DBLP QuAD Final test dataset for Scholarly QALD and 0.98 F1 score on the subset of the DBLP QuAD test dataset during the QA phase.

## 1. Introduction

Knowledge graphs, serve as an invaluable resource in the digital age. Their vastness and intricacies present challenges, especially when attempting to extract and understand relationships within. This complexity is intensified by the need for specific query languages to access these KGs. Recent advancements in the capabilities of language models (LMs) have transformed our understanding of natural language processing and its intersection with other domains of computer science. Language models like BERT, GPT-3, and Codex have set benchmarks for comprehending and producing diverse content, from natural language to complex code structures. A prime example is the PANGU[1] framework, which, with its discriminating abilities, enhances the alignment between LMs and user interfaces. However, we adopt simpler LM-based methods to set an efficient yet effective baseline.

While LMs have revolutionised various domains, their potential for interfacing KGs with natural language queries remain an area primed for exploration and refinement. The DBLP KG, a repository of scholarly data, can indeed serve as a valuable resource to elaborate on these emerging concepts.

By building upon the foundational work of existing models and methodologies, we introduce a novel two-phase strategy. The initial phase of relation extraction targets extracting immediate, one-hop relations using SPARQL, while identifying contextually pertinent labelled candidate pairs via heuristics, enhancing extraction precision. For the next phase of relation selection, we select a winning candidate from the extracted relationships using BERT's CLS token embeddings for similarity measurement, ultimately determining the strongest candidate based on cosine similarity values.[15]

## 2. Related Work

In the landscape of natural language processing (NLP), knowledge graph question answering (KGQA) systems have emerged as a promising solution. As the confluence of structured databases and intuitive query mechanisms, KGQA can facilitate the extraction of precise information in a user-friendly manner. At the core of many KGQA methodologies lies the focus on grounding the language models (LM) to real-world environments, which enables a robust question-answering mechanism over KGs.

Traditional approaches often leverage the Seq2Seq framework. This method employs a LM to generate a plan based on input statements [6][7]. A significant challenge for these methodologies is the lack of real-world grounding during the LM's initial training phase. This gap often results in complications in generating valid plans.

A recent study examined generalisation in question-answering on knowledge bases by looking beyond the conventional i.i.d. paradigm [3]. On a similar note, another study discussed ArcaneQA[5], which emphasised dynamic program induction and contextualized encoding for KGQA. This method can be considered precise, yet it is computationally intensive.

The current methods, while innovative, place a heavy dependency on extensive training data, and there's no assurance of maintaining consistent grammar or achieving absolute accuracy. Contrary to this trend, recent methodologies consider the LM primarily as a discriminator for valid plans suggested by an agent, driven through a structured search.

A fresh approach to grounding language models, emphasising their discriminative over generative capacities has been introduced with the PANGU framework[1]. This model, in KGQA applications, mirrors the bottom-up semantic parsing of the SmBoP model[11]. A recent study underlined KGQA challenges with unseen schema items, proposing a solution that synergizes a contrastive ranker with a tailored generation model. While the authors' claims suggest that the framework is KG-agnostic, it is difficult to adapt it to our use case – scholarly KGQA.

Despite efforts to adapt the initial SPARQL-to-S-expression conversion process within the PANGU codebase to our dataset, the subsequent code architecture remains intrinsically tailored to the Freebase KG. Adapting this framework to our scholarly KG requires extensive modification of the code to fit the distinct schema and query patterns, rendering the process non-trivial.

A noteworthy resource for scholarly KGQA is the DBLP scholarly knowledge graph (DBLP KG), an online bibliographic database focusing on major computer science publications. With its rich indexing of over 6.9 million publications from more than 3.3 million authors, the DBLP KG presents itself as a potent tool for KGQA research. This is exemplified by the introduction of DBLP-QuAD, a dataset specifically designed for question-answering over the DBLP KG[13]. With its inclusion of 10,000 question-answer pairs coupled with corresponding SPARQL queries, the DBLP-QuAD stands as the largest scholarly question-answering dataset to date.

## 3. Approach

Our approach is scaffolded into two main phases. A brief discussion about the code snippets used for different phases is detailed below.

```
Algorithm 1 Scholarly Knowledge Graph System
 1: function BERTOLOGYNAVIGATOR(Q : string) → Tuple[Entity, Relation]
 2:     # Step 1: Entity Extraction
 3:     entities ← ENTITYLINKINGAPI(Q)
 4:     E ← IDENTIFYRELEVANTENTITY(entities)
 5:     # Step 2: Relation Extraction
 6:     relations ← FETCHONEHOPRELATIONS(E, DBLP_KG)
 7:     # Step 3: Validation
 8:     valid_relations ← [relation for relation in relations if
    VALID_PAIR(relation)]
 9:     # Step 4: Determine Optimal Relation
10:     similarity_scores ← COMPUTESIMILARITY(valid_relations, Q)
11:     winning_candidate ← MAX(similarity_scores, key=similarity_scores.get)
12:     # Step 5: Output
13:     return (E, winning_candidate)
```

**Figure 1**: Algorithm for BERTologyNavigator.

The algorithm described is designed to answer queries about specific entities by using a knowledge graph system. Here is a breakdown of the algorithm and the system's workflow:

**Entity Extraction:** The system starts by extracting entities from the user's query using an entity linking API. It then pre-processes this information to determine the most relevant entity in the context of the query.

**Relation Extraction:** Once the relevant entity is identified, the system fetches 'one-hop' relations from DBLP KG.

**Validation:** The relations are then filtered through a validation process where labelled candidate pairs are evaluated against heuristics to ensure relevance to the query. These heuristics are based on filtering by relevance and may involve checking against specific identifiers like Wikidata, ORCID, or BibTeX keys.

**Determine Optimal Relation:** The valid relations are assessed for similarity to the query using BERT language model (LM) for cosine similarity. The system computes similarity scores between the query and the labelled candidate pairs.

**Output:** The system selects the winning candidate by identifying the labelled candidate pair with the highest similarity score. It checks if other candidates have the same predicate (P) and, if so, adds them to the result as well.

### 3.1. Phase 1: Relation Extraction

The initial phase focuses on two primary tasks: extracting one-hop relations and obtaining labelled candidate pairs, as one-hop relations typically give rise to simpler questions. Our approach handles simple questions with efficiency and accuracy.

**One-Hop Relations**: Leveraging the SPARQL query language and wrapper, immediate relations (one-hop) are extracted between entities from the DBLP KG. Through dual queries that fetch both incoming and outgoing relations and their respective labels, a comprehensive relationship mapping is acquired.

**Labelled Candidate Pairs**: We first use an entity linking API to identify relevant entities from the knowledge base. Next, the most contextually appropriate entity is selected for the given question. After entity selection, we retrieve possible labelled pairs associated with the identified entity and then filter them by several heuristics (such as entity type and property matching) for relevance against the question, ensuring pairs that are in no contextual alignment with the question are pruned.

These heuristics are designed to enhance the precision of the relationship extraction process. Notably, these conditions consider factors such as the presence of keywords in the question, the nature of predicates, and specific identifiers like "wikidata," "bibtex," "orcid," and more. Applying these heuristics, allow us to ensure that the selected pairs are relevant and contextually appropriate for answering user queries.

## 3.2. Phase 2: Relation Selection

Relation selection refines the relationships and entities identified in the prior phase by using BERT CLS Embeddings and a winning candidate selection mechanism.

**BERT Embeddings:** Utilising BERT's CLS token embeddings, a similarity measurement is implemented to discern the likelihood of a relationship between candidate pairs using cosine similarity. We also considered the dot product but went for the cosine similarity due to the computational cost.

**Winning Candidate Selection:** We identify the pair with the highest cosine similarity as the most promising candidate.

## 3.3. Implementation

The code for this paper is available at https://github.com/Shreyaar12/BERTology-Navigator
1. Relation Extraction: Using a Python based implementation, we rely on SPARQLWrapper to query the DBLP KG and retrieve one-hop relations and labels. Further, heuristic-based validation ensures the relevance of labelled pairs to the input question.
2. Winning Candidate Determination: Employing BERT's pre-trained models using the transformers library in Python, the `get_CLS _embedding` function retrieves the CLS token embeddings, and the cosine similarity between embeddings of candidate pairs and questions is computed to select the most promising relation.
3. Entity Linking: An API post-request is issued to fetch possible entities for a given question. The API endpoint integrates `t5-small` for label generation and `distmult` for embedding reranking to fetch potential entities relevant to the question context. A post-processing step further ensures the selection of the most relevant entity from the obtained candidates.

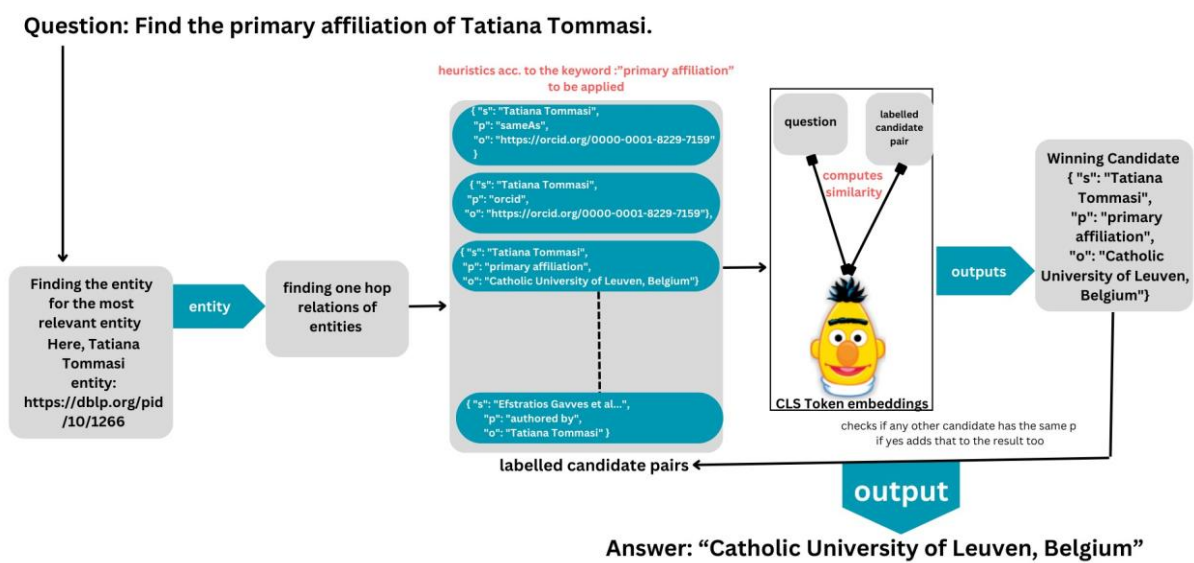

**Figure 2**: How a question is processed with BERTologyNavigator

## 4. Experimental Setup

### 4.1. Datasets

We utilise datasets derived from the DBLP QuAD for our experimental evaluation.

**A subset of the DBLP QuAD test dataset:** This is a curated set of 200 questions extracted from the DBLP QuAD test dataset. The questions were specifically chosen because they represent one-hop relations, enabling focused evaluation on this subset.

**Scholarly QALD Final Test Set:** Comprising 500 questions, this dataset is randomly drawn from DBLP QuAD. The final test set only includes the questions and their paraphrased versions, offering no additional context.

### 4.2. Implementation Details

Throughout the experiments, the entity linking in the model was done with the use of an entity linker tailored for the DBLP KG relying on the T5-small model[16][2].

## 5. Results and Discussion

The primary outcomes of BERTologyNavigator are depicted in Table 1. Initially, the system was assessed using a controlled dataset derived from the DBLP QuAD test dataset, emphasising questions with one-hop relations. This assessment yielded remarkable results, achieving a near perfect score in both entity linking and question answering, showcasing its capabilities in controlled settings. However, when subjected to the realistic dataset, the system presented an F1 score of 0.2175 for question answering, a marked reduction compared to its performance on the subset of the DBLP QuAD test dataset. While this score is notably lower, it offers a better reflection of BERTologyNavigator 's capabilities in handling complex real-world scenarios.

A notable observation was the significant drop in the question answering F1 score when heuristics were not applied, highlighting the system's reliance on such techniques for optimal performance.

**Table 1**
**Overall Results**

| Dataset | F1-Entity Linking | F1-Question Answering | F1-Question Answering without Heuristics |
| --- | --- | --- | --- |
| Scholarly QALD Final test set results | 0.6235 | 0.2175 | 0.0570 |
| Subset of the DBLP QuAD test dataset– 203 questions | 1.0 (using gold entities) | 0.9885 | 0.6341 |

A major factor contributing to the diminished F1 is the absence of multi-hop relation capabilities in the system. Multi-hop relations involve connecting entities through intermediate entities, which can be challenging to identify and resolve correctly. This complexity is especially apparent when the dataset contains paraphrased questions that may not directly mention all the required entities or relations.

### 5.1. Error Analysis

We conducted an error analysis on the BERTologyNavigator system using a sample of 203 questions from a subset of the DBLP QuAD test dataset. Our examination revealed that the system incorrectly performed entity linking on 44.3% (90 out of 203) of the questions, which significantly impacted the F1 score in both heuristic and non-heuristic scenarios.

With heuristics applied, we observed an issue that affected accuracy: despite correct entity recognition, 4 of the questions received incorrect answers due to the system not accounting for multiple URIs referring to the same entity. For instance, the question "What are the papers written by the person Ravi Netravali?" received an incorrect answer because the system failed to recognise that two different URIs—"https://dblp.org/rec/journals/corr/abs-2105-11118" and https://dblp.org/rec/conf/osdi/ThorpeQETHJWVNK21"—point to the same entity.

Without heuristics, the error rate increased to 83.7% (170 out of 203), with 44.3% of errors (90 out of 203) attributed to incorrect entity linking, 43.8% (89 out of 203) due to correct entities but incorrect answers, and an overlap where 39.9% (81 out of 203) of questions had both the wrong entity and the wrong answer.

For the final test set in the Scholarly QALD challenge, the system demonstrated strong performance on one-hop questions, which is reflected in the F1 score of 0.2175 for the question answering task. However, the system showed limited capability in handling multi-hop questions, contributing to the lower overall F1 score. A comprehensive analysis of the final test set was not feasible due to the absence of an answer key for this dataset.

## 6. Summary

The system has shown effectiveness and accuracy in identifying one-hop relations, particularly in a controlled environment. However, the challenge of real-world, complex questions, as seen in the Scholarly QALD challenge, underscores the necessity for enhancing the system's capabilities, specifically in handling multi-hop relations.

The progression towards multi-hop relation extraction and real-world applicability appears to be a logical next step, necessitating an intricate understanding of entities and their relational dynamics. This would entail an explorative journey into the realms of semantic understanding, relation prediction, and knowledge graph deepening.

## 7. Limitations

1. Lack of multi-hop Capability: The current system's design is focused solely on one-hop relations, restricting its applicability in more intricate datasets like the Scholarly QALD challenge.
2. Time accuracy: Our analysis indicates that while the system is capable of delivering accurate results, the current processing speed does not align with the requirements for real-time applications. Specifically, for one-hop questions, the system demonstrates an average response time of 7 seconds. However, the complexity of multi-hop questions significantly extends the processing time, with some queries taking upward of 10 minutes to resolve. Streamlining the system's efficiency is imperative to improve its suitability for real-time query processing and broaden its practical deployment scenarios.
3. Overfitting on the subset of the DBLP QuAD test dataset: The near perfect accuracy on the DBLP QuAD test dataset might indicate overfitting, where the system is tailored to a specific dataset.
4. Dependency on entity linkers: The accuracy and efficiency of the system are, in part, determined by the efficiency of the entity linkers it uses. If the entity linker performs poorly, the system's performance may be compromised.

Addressing these limitations in future iterations would be essential to make the system more resilient, versatile, and applicable across a wider range of datasets and scenarios.

## 8. Acknowledgement

We would like to extend our sincere gratitude to the University of Hamburg for hosting the first author during their research stay. We would also like to thank Debayan Banerjee for his valuable suggestions towards improving this work.